COVER SHEET

Title: **Impact of buckypaper on the mechanical properties and failure modes of composites**

Authors :
    Kartik Tripathi
    Mohamed H. Hamza
    Aditi Chattopadhyay
    Todd C. Henry
    Asha Hall

Paper Number: **004773**

**ABSTRACT**

Recently, there has been an interest in the incorporation of buckypaper (BP), or carbon nanotube (CNT) membranes, in composite laminates. Research has shown that using BP in contrast to nanotube doped resin enables the introduction of a higher CNT weight fraction which offers multiple benefits including higher piezo resistivity for health monitoring applications and enhanced mechanical response for structural applications. However, their impact on the deformation and failure mechanisms of composite laminates has not been investigated thoroughly. Understanding these issues experimentally would require a carefully executed test plan involving a multitude of design parameters such as BP geometry and placement, material anisotropy and variability, and laminate stacking sequence. Computational investigations can also be conducted to reduce the labor and cost associated with testing; however, for these results to be meaningful, high-fidelity physics-based simulation tools, accounting for scale-dependent variability, constitutive laws, and damage mechanisms and their evolution across the length scales, must be used. These methodologies are computationally intensive and their implementation in the analysis of complex heterogeneous structural systems can be prohibitive. This paper presents a deep learning (DL)-based surrogate model for studying the mechanical response of hybrid carbon fiber reinforced polymer (CFRP) composite laminates with BP interleaves under various mechanical loads. The surrogate model utilizes a long short-term memory (LSTM) architecture implemented within a DL framework and predicts the laminate global response for a given configuration, geometry, and loading condition. The DL framework training and cross-validation are performed via data acquisition from a series of three-point bend tests conducted through finite element analysis (FEA) and in-house experiments, respectively. The results show that the surrogate model is capable of predicting damage induced inelastic response without the computationally expensive task of solving nonlinear equations such as continuum damage mechanics and fracture mechanics-based equations. The model predictions show good agreement with FEA simulations and experimental results, where CFRP with two BP interleaves showed enhanced flexural strength and modulus over pristine samples. This enhancement can be attributed to the excellent crack retardation capabilities of CNTs, particularly in the interlaminar region. Finally, confocal microscopy images of experimentally tested specimens were analysed to interpret the predicted stress-strain response. Early damage initiation was observed in the 90° ply, resulting in the onset of nonlinearity and stiffness degradation. These micrographs also confirmed the role of BP in preventing through-thickness crack propagation.

Kartik Tripathi[1], Mohamed H. Hamza[1], Aditi Chattopadhyay[1], Todd C. Henry,[2] Asha Hall[2],

[1]School for Engineering of Matter, Transport, and Energy, Arizona State University, Tempe, AZ, USA

[2]DEVCOM Army Research Laboratory, Aberdeen Proving Ground, MD, USA

# 1. INTRODUCTION

There has been significant interest in the development and utilization of novel materials for improved performance in applications ranging from aerospace and electronics to energy and healthcare. Carbon Nanotubes (CNTs) have emerged as an exciting and promising material with vast potential for a wide array of applications due to their extremely high tensile strength (100 GPa) and stiffness (1 TPa) properties. In addition to their superior mechanical properties, CNTs also possess excellent thermal (~3500 W/m/K) and electrical (~$10^5$ S/m) conductivities making them suitable for use in de-icing and lightning strike protection applications in the aerospace industry [1,2]. The cylindrical structure of the CNTs, formed by rolled-up graphene sheets, offers excellent mechanical stability and strength while maintaining flexibility and resilience. Despite the exceptional material properties exhibited by CNTs, harnessing their full potential for structural applications presents a formidable challenge [3]. Researchers have explored the potential of CNT-doped resins as a host matrix to achieve desired material properties. However, a significant drawback associated with this approach is linked to the tendency of CNTs to agglomerate due to Van der Walls forces, leading to property variability at the macro-scale. Additionally, there is a limitation on the weight fraction of CNTs that can be effectively doped, which subsequently restricts the full potential utilization of CNTs in such systems.

Thin membranes composed of CNTs interconnected in a mesh-like network, also known as buckypaper (BP), offer a viable solution by enabling the incorporation of high weight fractions of CNTs and ensuring a uniform dispersion throughout the composite structure [4]. This approach circumvents the heterogeneous properties observed at the macro-scale and expands the potential applications of BPs to a variety of applications which include water purification, gas/vapor sensing, strain sensing, fire retardant coatings, artificial muscles, electromagnetic interference shielding and self-heating hybrid composites [5]. The porous nature of BP facilitates a strong interface bond with neighboring plies making it compatible for integration into laminate structures. Numerous researchers have successfully fabricated and characterized high-performance BP composites, specifically designed to address challenges in contemporary aerospace structures [5,6]. The presence of BP has been shown to significantly enhance the interfacial strength, which has become a topic of great interest among researchers. Recent studies have demonstrated an approximate ~ 45.9% improvement in the mode II fracture toughness ($G_{IIc}$), a critical factor influencing delamination behavior, through the incorporation of BP [7]. However, there has been little effort to understand damage initiation and evolution mechanisms in such composites.

The objective of this paper is to formulate a deep learning-based (DL) response surface model for comprehensive understanding of the relationship between BP placement and the mechanical response of carbon fiber reinforced polymer (CFRP) laminates under flexural loading. The methodology is validated using finite element analysis (FEA) simulations and experiments. Microstructural analysis of pristine and failed BP-integrated hybrid specimens was conducted to provide insights into the predicted response. The DL-based framework can be further extended to model a wide range of loading environment, enabling optimization of the configuration for desired performance while maintaining computational efficiency.

## 2. METHODOLOGY

A DL-based surrogate model is developed to emulate the response of hybrid laminates, with varying number of plies, ply orientations, and BP stacking sequence, and applied load. Data acquisition is conducted for DL training and testing. A total of 2000 load-displacement datasets were generated by simulating three-point bend tests using FEA. Various hybrid laminate configurations were simulated to create a database representing a wide range of design possibilities [8]. BP films are introduced at different locations within the CFRP laminate to evaluate its impact on key performance indicators such as strength and stiffness. The FEA results served as the foundation for training the surrogate model. To ensure the reliability and accuracy of the model, in-house experiments were performed, and their results were compared with the response surface model for model cross-validation, and for further analysis of the hybrid laminates mechanical response and failure mechanisms. Miralon S-T01AVB-12 sheets sourced from Nanocomp Technologies were used as BP interleaves between CFRP plies. These BP films possess a randomly interconnected network of CNTs held together by Van der Walls forces. Scanning electron microscope (SEM) was used to illustrate the resin infused BP morphology at the hybrid laminate ply interface as shown in Figure 1. The interconnected structure allows for efficient stress transfer and crack propagation resistance, making it well-suited for enhancing the performance of composite materials. As there is no inherent directional preference observed, the BP films are assumed as isotropic in nature. The mechanical properties of the BP were obtained from both the manufacturer and recent literature [9,10] and are presented in Table I where $E$, $v$, $\sigma_{ut}$, $\tau_s$, and $\rho$ represent the elastic modulus, Poisson's ratio, normal strength, shear strength, and density of the BP film, respectively. The composite material system includes unidirectional (UD) plies made from IM7 carbon fiber and 8552 epoxy resin, with their mechanical properties provided in Table II from technical datasheets and recent literature [11-13]. $E_{11}$, $G_{12}$, $G_{13}$, $v_{12}$, and $v_{13}$ denote the elastic modulus, shear modulus and Poisson's ratio of the ply along the longitudinal direction, whereas $E_{22}$, $E_{33}$, $G_{23}$, and $v_{23}$ denote the mechanical properties in the transverse direction. These material properties correspond to a 60% fiber volume fraction denoted by $V_f$. In this section, we present a comprehensive account of the numerical model and the setup procedures employed in the study.

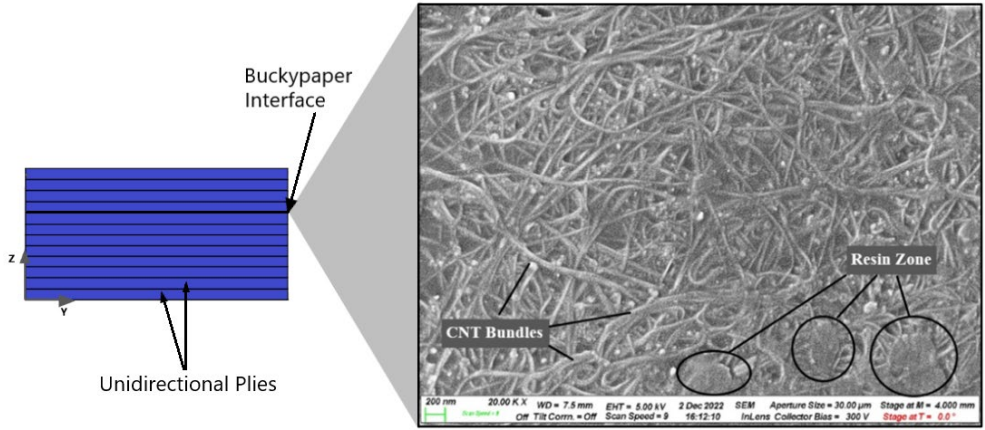

**Figure 1**. Hybrid Laminate Schematic with BP micrographs using SEM.

Table I. Mechanical properties of resin infused BP film [9,10].

| $E$ (GPa) | $v$ | $\sigma_{ut}$ (MPa) | $\tau_s$ (MPa) | $\rho$ (g/cc) |
|---|---|---|---|---|
| 5.8 | 0.33 | 50.8 | 15 | 0.65 |

Table II. Mechanical properties of 0° IM7/8552 lamina [11-13].

| $E_{11}$ (GPa) | $E_{22}, E_{33}$ (GPa) | $v_{12}, v_{13}$ | $v_{23}$ | $G_{12}, G_{13}$ (GPa) | $G_{23}$ (GPa) | $V_f$ |
|---|---|---|---|---|---|---|
| 162.1 | 8.97 | 0.32 | 0.35 | 4.69 | 3.92 | 0.6 |

## 2.1 Numerical Analysis

The load response of the laminates was predicted by employing a progressive damage model, encompassing both the intra-laminar and inter-laminar regions. The interlaminar region of the laminate was modeled as a cohesive zone (CZ), where each interface within the laminate exhibited cohesive interaction behavior. To ensure a realistic representation of the contact behavior, a surface-to-surface contact formulation with small sliding was chosen. This approach, compared to finite sliding, is deemed more appropriate as it allows for small relative displacements and rotations while maintaining contact integrity. Depending on the interface type, each ply interface was assigned either a BP or resin CZ property. Table III provides a summary of the CZ properties for both types of interfaces [4,7-9]. $K$ and $\tau$ represent the stiffness and strength of the CZ respectively with subscripts $n$, $s_1$, $s_2$ denoting the normal and shear directions respectively. $G_{Ic}, G_{IIc}$, and $G_{IIIc}$ represent the interlaminar fracture toughness in the three failure modes. The damage initiation within the CZ is modeled through quadratic nominal stress criterion and its evolution is governed by Benzeggagh-Kenane (BK) criterion with BK constant η given in Table III [14]. It is worth noting that the BP interface offers significant improvements, particularly in terms of mode II fracture toughness, when compared to the neat resin interface. This plays a crucial role in bolstering the overall strength and integrity of the interface, in the case of hybrid laminates.

Table III. Properties of CZ interactions for resin and BP [4,7-9].

|  | Resin | BP |
|---|---|---|
| $K_n$ (MPa/mm) | 2500 | 5800 |
| $K_{s2}, K_{s2}$ (MPa/mm) | 864 | 2071 |
| $\tau_n$ (MPa) | 37 | 50 |
| $\tau_{s1}, \tau_{s2}$ (MPa) | 11 | 16 |
| $G_{Ic}$ (MPa mm) | 0.24 | 0.32 |
| $G_{IIc}, G_{IIIc}$ (MPa mm) | 0.775 | 1.18 |
| η | 2.67 ||

3D Hashin failure criteria that account for the strengths of the lamina in various directions and their interactions were used to model intra-laminar damage and failure of the hybrid laminate. These failure criteria, shown in Eqs. 1-6, include fiber tension/compression, in-plane matrix cracking/crushing, and out-of-plane matrix cracking/crushing [14]. $e_{ft}$, $e_{fc}$, $e_{imt}$, $e_{imc}$, $e_{omt}$, $e_{omc}$ denotes the indices of fiber, matrix in-plane and out-of-plane failures under tension and compression loading respectively. $\varepsilon_i$ ($i$ = 1, 2, 3) are the engineering strains along the principal material axes at the given time instance and $C_{ij}$ ($i, j$ = 1, 2, 3; $i \neq j$) are the elements of the stiffness matrix. $X$ denotes the failure strengths of the ply along the fiber direction, while $Y$ and $Z$ denote the transverse failure strengths. The subscripts $t$ and $c$ correspond to tensile and compressive strengths, respectively. Additionally, $S$ is the shear strength and assumed to be uniform across all planes. The failure loads of an IM7/8552 UD lamina are summarized in Table IV [15,16].

Fiber Tensile Failure, ($\varepsilon_1 > 0$):
$$e_{ft}^2 = \left(\frac{C_{11}\varepsilon_1}{X_t}\right)^2 + \left(\frac{C_{66}\gamma_{12}}{S_{12}}\right)^2 + \left(\frac{C_{55}\gamma_{13}}{S_{13}}\right)^2 \geq 1 \quad (1)$$

Fiber Compressive Failure, ($\varepsilon_1 < 0$):
$$e_{fc}^2 = \left(\frac{C_{11}\varepsilon_1}{X_c}\right)^2 \geq 1 \quad (2)$$

In-plane Matrix Cracking, ($\varepsilon_2 > 0$):
$$e_{imt}^2 = \left(\frac{C_{22}\varepsilon_2}{Y_t}\right)^2 + \left(\frac{C_{66}\gamma_{12}}{S_{12}}\right)^2 + \left(\frac{C_{44}\gamma_{23}}{S_{23}}\right)^2 \geq 1 \quad (3)$$

In-plane Matrix Crushing, ($\varepsilon_2 < 0$):
$$e_{imc}^2 = \left(\frac{C_{22}\varepsilon_2}{Y_c}\right)^2 + \left(\frac{C_{66}\gamma_{12}}{S_{12}}\right)^2 + \left(\frac{C_{44}\gamma_{23}}{S_{23}}\right)^2 \geq 1 \quad (4)$$

Out of plane Matrix Cracking, ($\varepsilon_3 > 0$):
$$e_{omt}^2 = \left(\frac{C_{33}\varepsilon_3}{Z_t}\right)^2 + \left(\frac{C_{55}\gamma_{13}}{S_{13}}\right)^2 + \left(\frac{C_{44}\gamma_{23}}{S_{23}}\right)^2 \geq 1 \quad (5)$$

Out of plane Matrix Crushing, ($\varepsilon_3 < 0$):
$$e_{omc}^2 = \left(\frac{C_{33}\varepsilon_3}{Z_c}\right)^2 + \left(\frac{C_{55}\gamma_{13}}{S_{13}}\right)^2 + \left(\frac{C_{44}\gamma_{23}}{S_{23}}\right)^2 \geq 1 \quad (6)$$

Table IV. 0° IM7/8552 Lamina failure properties [15,16].

| $X_t$ (MPa) | $X_c$ (MPa) | $Y_t = Z_t$ (MPa) | $Y_c = Z_c$ (MPa) | $S_{12} = S_{13} = S_{23}$ (MPa) | $G_f$ (N/mm) | $G_m$ (N/mm) |
|---|---|---|---|---|---|---|
| 2558 | 1732 | 64 | 285 | 91 | 106 | 0.263 |

An exponential progressive damage evolution parameter denoted as $d_k$ is given by Eq. 7, where the subscript $k$ denotes the associated failure mode. At any specific point in time, the damage state of an element can be characterized by three distinct damage parameters, each corresponding to a specific mode of failure – fiber failure ($d_f$), in-plane matrix failure ($d_{im}$), and out of plane matrix failure ($d_{om}$) for either under tension or compression loading. $\sigma_n$ and $\varepsilon_n$ denote the corresponding engineering normal stress and strain components of the damage state at the given instant along the principal material axes. $G_p$ is the fracture energy of either fiber ($p = f$) or matrix ($p = m$) depending on the associated mode of failure. $G_f$ and $G_m$, given in Table IV, are vital in accurately capturing the intricacies of intra-laminar damage evolution. Finally, the characteristic length of the mesh element is denoted by $L_c$ which helps in reducing the influence of mesh size on energy dissipation, ensuring reliable results [17].

$$d_k = 1 - \frac{1}{e_k} exp\left(\frac{\sigma_n \varepsilon_n (1 - e_k) L_c}{G_p}\right) \quad (7)$$

To effectively account for the degradation of the transversely isotropic stiffness [18], a user-defined material subroutine (UMAT) was developed, providing a robust framework for accurately capturing the degradation behavior within the material. Using the damage parameters $d_k$ stored as state dependent variables for each failure mode, the stiffness elements undergo degradation according to Eq. 8, resulting in an updated term denoted as $C_{ij}^d$ as follows:

$$\begin{aligned}
C_{11}^d &= (1 - d_f)^2 C_{11}, & C_{12}^d &= (1 - d_f)(1 - d_{im}) C_{12}, \\
C_{22}^d &= (1 - d_{im})^2 C_{22}, & C_{13}^d &= (1 - d_f)(1 - d_{om}) C_{13}, \\
C_{33}^d &= (1 - d_{om})^2 C_{33}, & C_{23}^d &= (1 - d_{im})(1 - d_{om}) C_{23}, \\
C_{44}^d &= (1 - d_{im})(1 - d_{om}) C_{44}, & C_{55}^d &= (1 - d_f)(1 - d_{om}) C_{55}, \\
C_{66}^d &= (1 - d_f)(1 - d_{im}) C_{66}
\end{aligned} \quad (8)$$

The methodology is implemented numerically using FEA via Abaqus. The construction of the FEA model is depicted in Figure 2, and the detailed geometry information can be found in Table V. During the simulations, both the reaction force and displacement components along the Z-axis were recorded at a reference point rigidly constrained with the load line at the span center. To emulate the experimental conditions of the 3-point bending test, the support lines at the bottom of the laminate were pinned, hence restricting any displacement. Furthermore, the reference point was subjected to a predefined velocity boundary condition in the negative Z-direction, as depicted in Figure 2(a). Since the structure was relatively simple, a structured mesh with four elements through the ply thickness and an overall density of ~3 elements/$mm^3$ was implemented for each ply as shown in Figure 2(b). For numerical stability and computational efficiency, trilinear hexahedral 8 node "brick element" with reduced integration (C3D8R) and enhanced hourglass stiffness were chosen for the simulations.

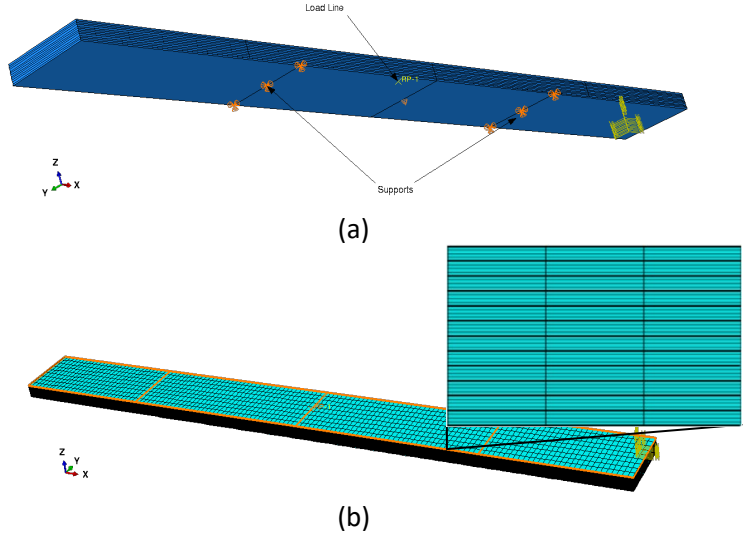

(a)

(b)

**Figure 2.** FEA test setup, **(a)** specimen geometry with boundary conditions, **(b)** meshed laminate with enlarged view of refined through thickness region.

**Table V.** Specimen ply and geometry details.

| Number of Plies | Ply Thickness (mm) | Span Length (mm) | Specimen Geometry | | |
|---|---|---|---|---|---|
| | | | Length (mm) | Width (mm) | Thickness (mm) |
| 12 | 0.25 | 60 & 80 | 140 | 14 | 3 |

To automate the data generation for the surrogate model, python scripts were developed to generate 2000 input files. The datasets were generated by varying the baseline laminate configuration and geometry, BP placement, and the CZ properties. 11 commonly used symmetric laminate configurations such as [90, 0, 45, -45, 0, 90]$_s$, [(0, 90, 0)$_2$]$_s$ etc. were used and the CZ strength and fracture toughness were uniformly varied within a small range to account for material properties variability as reported in literature. To improve the computational efficiency for training data generation, a coarse mesh with a density of ~1 element/$mm^3$ was used for each ply. For each simulation, the field history output for the reference point was exported at a fixed time step of 1.5 seconds. A total of 1000 output files were generated for each span length of 60 mm and 80 mm. These files, containing the required data, were utilized as the training data set for the surrogate model.

*2.2 Experiments and Model Validation*

The three-point bending experimental tests were conducted to assess the improvements in flexural strength and stiffness of hybrid laminates and compare the findings with the predictions of the FEA and surrogate model. The experimental study involved testing a total of four laminate configurations as shown in Figure 3 following the standard test method recommended by ASTM (D-7264) [19]. Table VI further summarizes the stacking sequence for the four laminate configurations considered for experimental tests. To cross-validate the numerical and DL models predicted global responses, baseline (12/0) and hybrid (12/2) laminate configurations were fabricated and tested. Meanwhile, hybrid (12/3) and (12/4) laminates were used to further analyze the mechanical response of incorporating BP in the laminate with varying position (symmetric to laminate midplane) and number of BP layers.

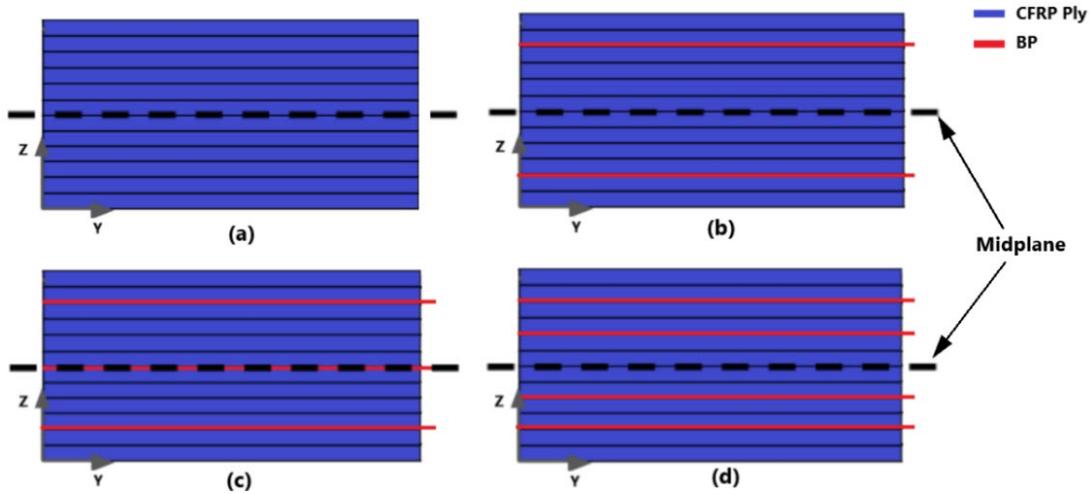

**Figure 3.** The four hybrid laminate configurations illustrating the BP placement, (a) baseline (12/0), (b) (12/2), (c) (12/3), and (d) (12/4) laminates.

**Table VI.** Laminate configurations for experimental tests.

| Baseline (12/0) | – [90, 0, | +45, -45, | 0, 90]$_s$ |
|---|---|---|---|
| 2 BP interleaves (12/2) | – [90, 0, BP, | +45, -45, | 0, 90]$_s$ |
| 3 BP interleaves (12/3) | – [90, 0, BP, | +45, -45, | 0, 90, $\overline{BP}$]$_s$ |
| 4 BP interleaves (12/4) | – [90, 0, BP, | +45, -45, BP, | 0, 90]$_s$ |

For the experimentation, the given laminates of dimension 300×300 mm were fabricated by means of a hand layup. To prevent any resin leakage onto the equipment surface, a Teflon film was applied to cover the stack, which was then sandwiched between two metal plates. Subsequently, the entire assembly was placed into a hot press. The laminate underwent a curing process at 70°C and 2 MPa of pressure for a total duration of 5 hours. After the curing process, test coupons sized at 140 x 14 mm were cut, sanded, and subjected to three-point bend tests to evaluate their mechanical response. The three-point bend tests were performed using a 3 kN dynamic load frame equipped with a suitable fixture, as depicted in Figure 4.

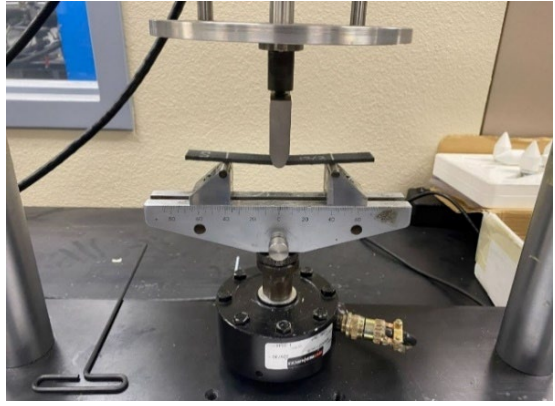

**Figure 4.** Experimental setup for three-point bending test.

A support span of 80 mm was employed, and a displacement loading was applied at the span center at a constant rate of 2 mm/min. The raw data obtained from the load frame, FEA and surrogate model simulations was used for determining the stress-strain curves, and flexural strength according to Eqs. 9&10. Here, the engineering stress ($\sigma_b$) and strain ($\varepsilon$) along the laminate X-axis at the span center of inner/outer-most ply depend on reaction force ($P$), deflection $\delta$, and specimen length ($L$), width ($b$), and thickness ($t$) as follows:

$$\sigma_b = \frac{3PL}{2bt^2} \quad (9)$$

$$\varepsilon = \frac{6\delta t}{L^2} \quad (10)$$

## 2.3 Surrogate Model Development

Recurrent neural network (RNN) has been widely implemented to solve data dependencies problems such as natural language processing, flight dynamics and constitutive models [20-22]. Usually, during RNN training the learning parameters such as the weight matrices and bias vectors are being updated through automatic differentiation, where backpropagation is applied through layers as well as through history variables steps (i.e., backpropagation through strain increments history). However, for complex DL architectures this can lead to vanishing gradients and ultimately training terminates without loss function convergence. Hence, variations from the standard RNN architecture such as long short-term memory (LSTM) and gated recurrent unit have been developed to address this issue [23,24]. LSTM is capable of learning long-term dependencies while dynamically modifying the history variables to determine the percentage of previous time steps that can contribute to the prediction at the current time step.

The schematic of the proposed surrogate model NN architecture is shown in Figure 5. The LSTM architecture can be embedded into a DL framework to learn complex mapping between input and output tensors. Convolutional neural network (CNN) can enhance the accuracy of LSTM training accuracy especially in case of multi-dimensional input such as the load curve and laminate configurations. The input to the model is sliced into sequential and non-sequential data parts, the load curve in terms of applied displacement ($d_t$) represents the sequential part, which is fed into two LSTM layers, while skip connection is applied to bypass the LSTM layers and connect the laminate stacking sequence including BP locations as well as geometry to CNN layers. Analogous to 2D CNN, which is widely applied in image processing techniques, 1D CNN is used to highlight the main contributing non-sequential input data to the laminate constitutive response. CNN and LSTM output distributions are concatenated and passed through fully connected layers (FCLs), while applying batch normalization to avoid internal covariate shift and possible convergence issues [25]. FCLs are finally connected to the output linear layer to predict the load-displacement curve of the laminate. The DL-based surrogate model is trained, tested, and cross-validated on NVIDIA TITAN RTX GPU using the TensorFlow 2.11.0 package and Python 3.7.2. Adam optimizer [26] is used to perform adaptive moment stochastic gradient descent with learning rate equal to $10^{-3}$ and 64 mini-batch size, while root mean squared error (RMSE) is selected as the loss function for the ground truth and predicted load-displacement curve. Finally, the model is trained on 90% and tested on 10% of the numerical simulations (FEA) data. The training dataset for surrogate model training obtained from FEA automation was limited to only pristine and hybrid laminates with two embedded BP layers at arbitrary locations while maintaining laminate overall symmetry. This selection was made due to the model's inability to effectively capture the influence of weak zones which are caused by addition of excessive BP layers within the laminate, this was further discussed in Section 3.2. Upon successful training, the model is cross-validated on the experimental data for (12/0) and (12/2) hybrid laminate configurations.

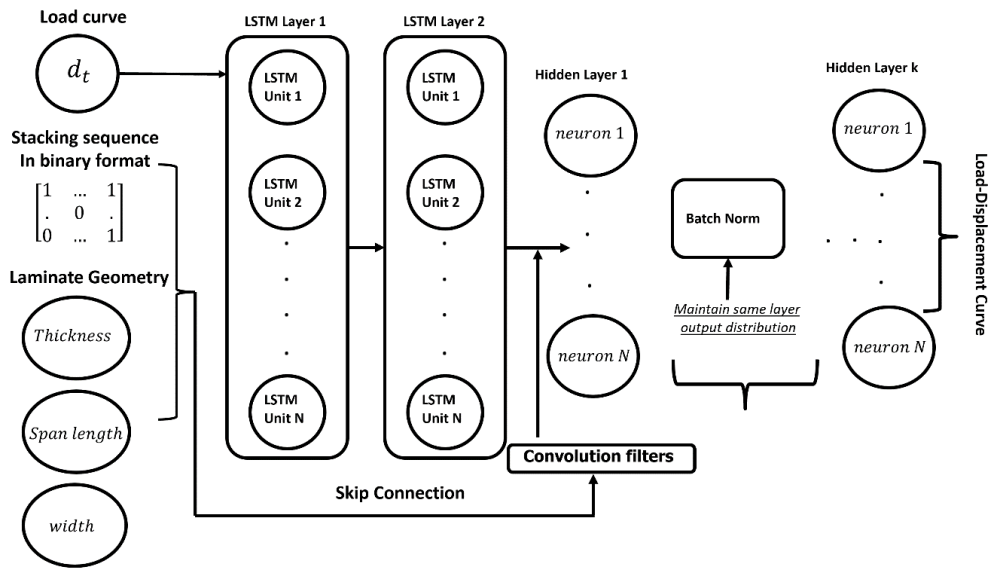

**Figure 5.** DL-based surrogate model architecture which consists of LSTM coupled with convolutions and FCLs.

## 3 RESULTS AND DISCUSSION

### 3.1 Surrogate Model

The generated load-displacement curve from the FEA is compared with experimental results for model validation. The stress-strain comparison for the baseline (12/0) and hybrid (12/2) laminate configurations is presented in Figure 6. In the FEA results, a slight degradation in stiffness is observed around 0.5% strain. This degradation is primarily attributed to the occurrence of transverse tensile failure in the outermost ply. As the tensile stress in the transverse direction rises, the damage parameter associated with the in-plane matrix cracking $d_{im}$ approaches 1 indicating ply failure as shown in Figure 6. This transverse tensile failure leads to a reduction in stiffness in the FEA results. The ply damage parameters evolution is approximately same for pristine and hybrid laminates simulations, meanwhile the BP CZ properties and its damage evolution are the key parameters in hybrid laminates deformation behavior. The discrepancies in the FEA and experimental results can be attributed to several factors, such as manufacturing related defects and uncertainties in microstructure morphology and architectural parameters in the test specimens.

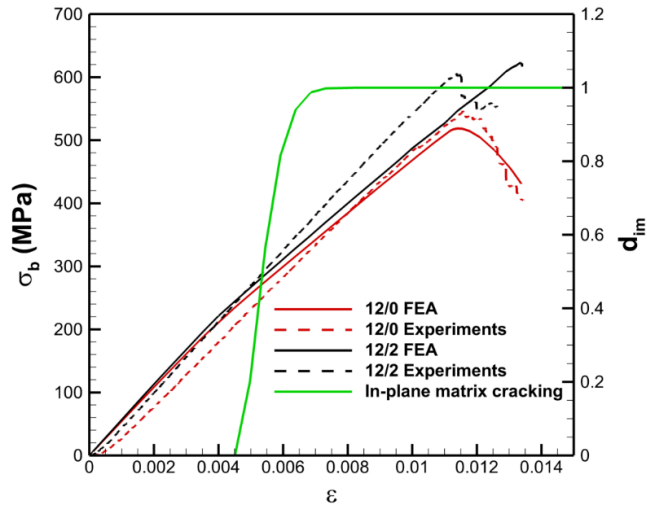

**Figure 6.** Stress-strain and in-plane matrix cracking damage evolution results for (12/0) and (12/2) laminates obtained from FEA and experiments.

The evolution of damage near the middle of the span length of the (12/0) laminate is shown in Figure 7. The damage parameters corresponding to in-plane matrix failure ($d_{im}$) and fiber failure ($d_f$) are depicted. In the initial stages of loading, no signs of damage are observed. However, as the displacement reaches a value of 2 mm, failure initiation is observed as indicated by the $d_{im}$ contours in Figure 7(a). The 90° ply located farthest from the neutral axis is the first to fail due to its relatively low tensile strength in the transverse direction. As the load is increased, at approximately 4 mm displacement, further damage evolution is observed, leading to compressive failure of the 0° ply, as shown in Figure 7(b).

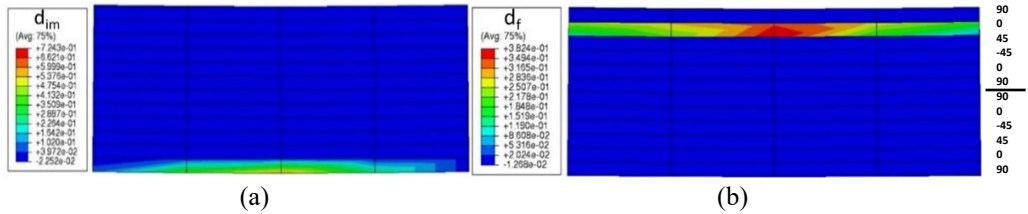

(a)              (b)

**Figure 7.** Damage evolution near the span center of the (12/0) laminate, (a) transverse tensile failure of the 90º ply at 2 mm displacement, (b) compressive fiber failure of the 0º ply at 4 mm displacement.

The DL surrogate model is trained on mini batches till the RMSE training and testing losses saturate indicating a fully convergent surrogate model as indicated in Figure 8(a). It is to be noted that the testing loss decreases with training steps and saturates at a lower level compared to the training loss due to kernel and bias regularizations which are only applied during training. Upon successful training the model is tested on various laminate configurations which represent the typical symmetric hybrid CFRP response under different stacking sequences and two BP various placements. Figure 8(b) shows a sample of the load-deflection curve for FEA and DL results, where the DL model captures the hybrid laminate effective stiffness and onset of nonlinearities resulting from stiffness degradation as indicated in Eq. 8. Additionally, the model captures the softening response due to delamination resulting from CZ damage initiation and evolution under loading. The DL model response slightly deviates from the FEA ground truth response upon delamination due to increasing nonlinearities accompanied with significant strain energy dissipation. Further improvements in the DL model can be attained through constraining the loss function with consistent tangent modulus at each loading step.

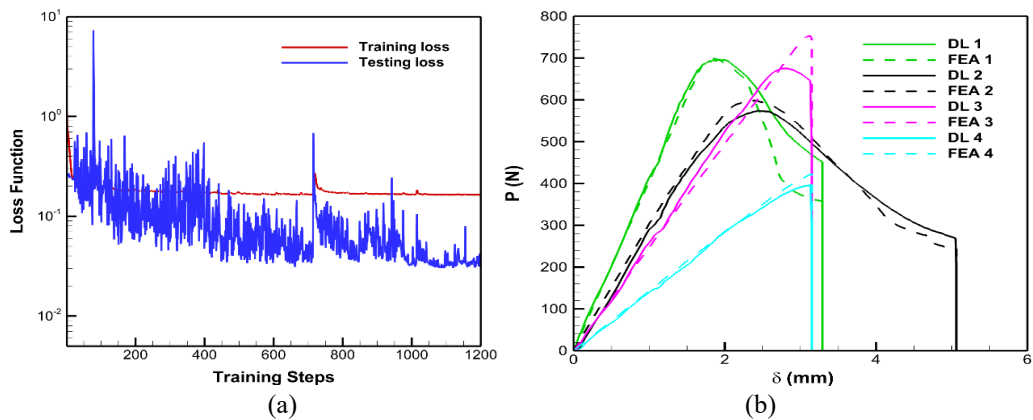

(a)              (b)

**Figure 8.** DL surrogate model results, (a) RMSE training and testing losses variation with training steps, (b) predicted load vs displacement response for different configurations and CZ properties.

The surrogate model is cross-validated with three-point bending tests; the stress response is compared with the experimental results for (12/0) and (12/2) CFRP laminates where a given applied strain rate and laminate configuration is fed into the model and as shown in Figure 9. Analogous to Figure 8, the model captures the overall response including damage initiation, evolution and the strain at which delamination occurs. The model underestimates the flexural strength, this can be attributed to the fact that the model is trained on FEA results with coarser mesh as discussed in Section 2.2. However, the model shows good agreement with experimental results in terms of flexural stiffness, and the degradation due to the 90° ply damage evolution which are crucial aspects for hybrid CFRP laminate design. Finally, the computational efficiency of the DL surrogate model is compared against the FEA simulations at mesh density of ~1 element/$mm^3$. For fair comparison, both FEA and proposed model are run on high throughput computing CPU nodes. The surrogate model showed an average 3.5 order of magnitude enhancement in computational cost, such an improvement in efficiency has a drastic impact on the structural complexity and the domain size which can be analyzed.

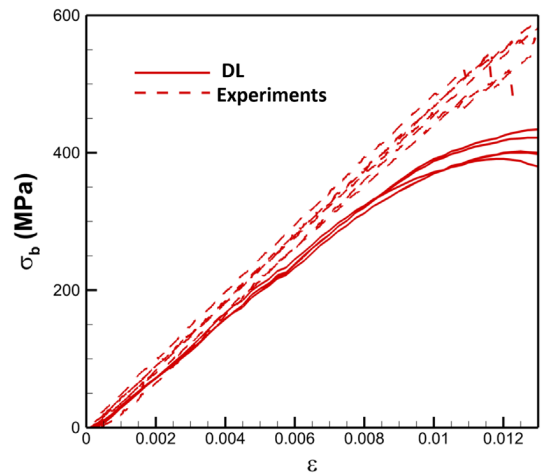

**Figure 9.** DL surrogate model and experimental stress-strain results for (12/0) and (12/2) laminates.

*3.2 Experimental Analysis*

A comprehensive analysis was conducted on seven specimens for each of the four configurations summarized in Table VI. The results indicate that hybrid laminates with a BP interleaf exhibited superior performance in the three-point bending tests. Figure 10 illustrates the average stress-strain response of each laminate configuration. Notably, the (12/2) configuration demonstrated the highest strength and stiffness among all the tested configurations. As compared to the baseline (12/0) configuration, the (12/2) configuration showed about 15% increase in flexural strength and a 10% increase in the stiffness. This improvement can be credited to the uniform network of interconnected high aspect ratio CNTs, which facilitated improved load transfer and interfacial strength. Another interesting observation, as depicted in Figure 10, is that the addition of BP layers beyond a certain threshold in the laminate results in degradation of mechanical properties. This degradation can be attributed to introduction of porosity at multiple laminate locations, where eventually microcracks can propagate along the BP and resin interface and then traverse through ply at potential weak spots, leading to a decline in the overall mechanical performance. Hence, the porosity of the BP, location along the stacking sequence and the ability of the resin to adequately infuse it are crucial factors that significantly influence the effective properties of the composite material.

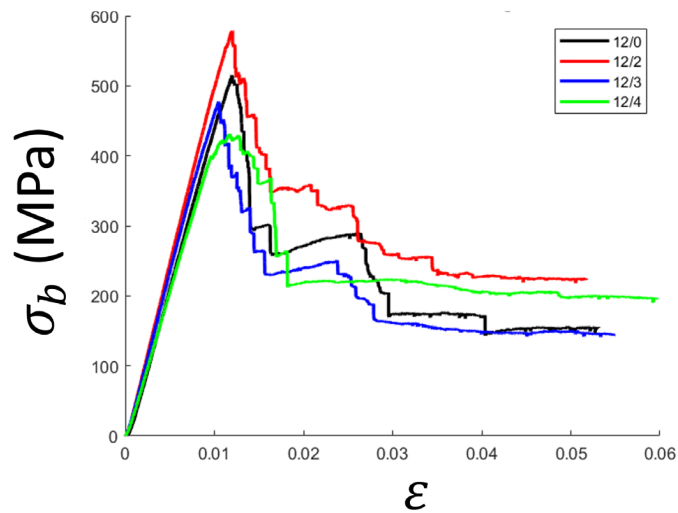

**Figure 10.** Average stress-strain response of pristine and hybrid laminates.

Confocal microscope images of damaged samples under different displacements, specifically 4 mm and 12 mm are presented in Figure 11(a) and (b) respectively. Among the four configurations examined, the sample with two BPs exhibits the slowest damage evolution. An enlarged view of the midplane of the damaged samples is depicted in Figure 11(c), revealing the hindrance in through-ply crack propagation due to the presence of the BP interleaf. In contrast, in configurations without a BP at the midplane, the crack propagates directly through the ply thickness. This behavior is likely due to crack tip blunting at the BP interface. As the crack comes in contact with the BP, localized deformation and energy absorption within the CNT network causes the crack to deflect. The presence of the BP and the associated CNT network effectively hinder the crack propagation, leading to improved fracture resistance and delayed structural failure.

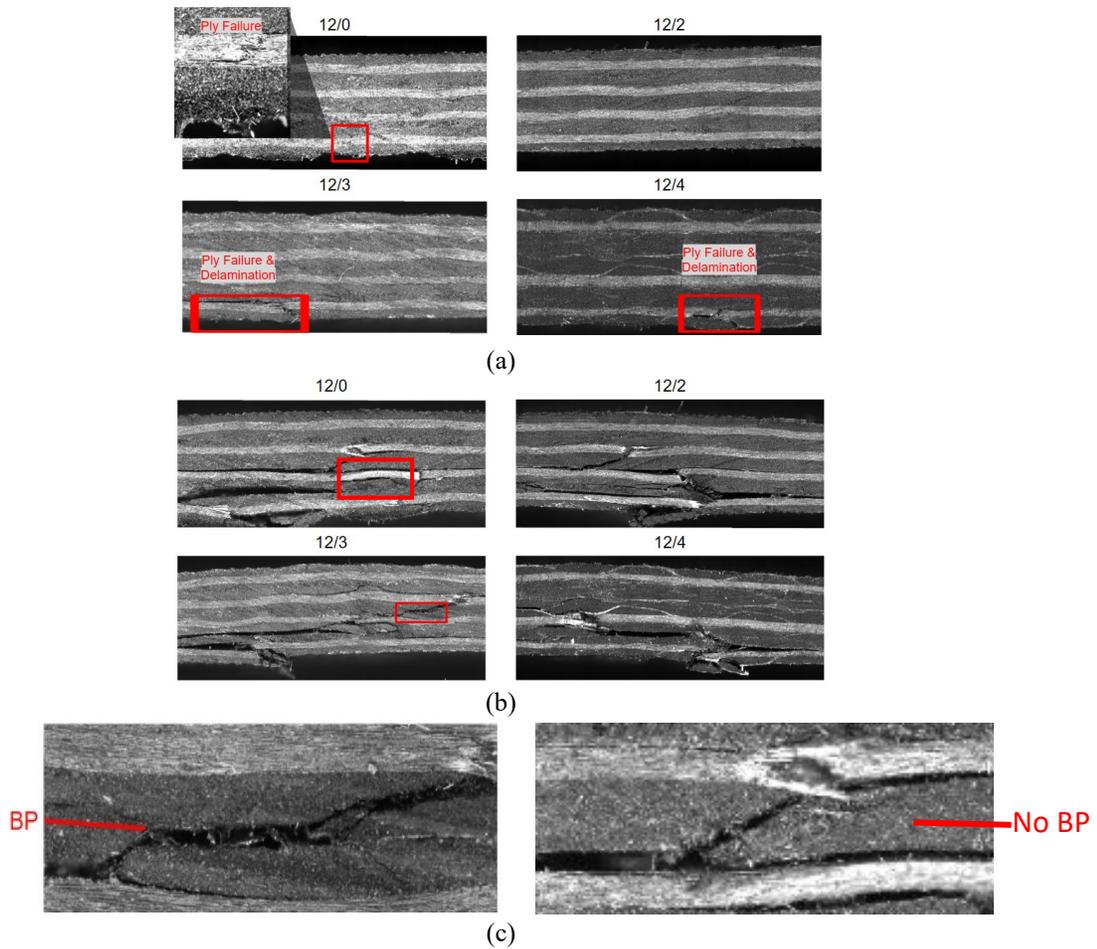

**Figure 11.** Confocal microscope images of damaged specimens, (a) at 4 mm displacement, (b) at 12 mm displacement, (c) BP layer delaying through thickness crack propagation as compared to pristine specimen.

## 4 CONCLUSIONS

This study aimed to develop a surrogate model to emulate hybrid CFRPs constitutive behavior based on load-displacement data obtained from numerical simulations of multiple laminate configurations. The experimental results were utilized to validate the surrogate model which serves as a response surface tool, enabling a comprehensive understanding of the response of hybrid CFRPs configurations, while maintaining computational efficiency. The incorporation of BP in the laminate demonstrated a significant improvement on the overall strength through improving the interlaminar region. However, excessive addition of BP resulted in the formation of weak spots, which ultimately lead to a reduction in laminate strength. The developed tool provides valuable insights into the behavior of different configurations, making it beneficial for studying and optimizing laminate designs for specific applications.

# ACKNOWLEDGMENTS

Research was sponsored by the Army Research Laboratory and was accomplished under Cooperative Agreement W911NF-17-2- 0207. The views and conclusions contained in this document are those of the authors and should not be interpreted as representing the official policies, either expressed or implied, of the Army Research Laboratory or the U.S. Government. The U.S. Government is authorized to reproduce and distribute reprints for Government purposes notwithstanding any copyright notation herein.

# REFERENCES

[1] Iijima, S. (1991). "Helical microtubules of graphitic carbon." Nature, 354(6348), 56–58
[2] Eatemadi A, Daraee H, Karimkhanloo H, Kouhi M, Zarghami N, Akbarzadeh A, Abasi M, Hanifehpour Y, Joo SW. "Carbon nanotubes: properties, synthesis, purification, and medical applications" Nanoscale Res Lett. 2014 Aug 13;9(1):393
[3] Li, Jianbin, Zhang, Zhifang, Fu, Jiyang, Liang, Zhihong and Ramakrishnan, Karthik Ram. "Mechanical properties and structural health monitoring performance of carbon nanotube modified FRP composites: A review" Nanotechnology Reviews, vol. 10, no. 1, 2021, pp. 1438-1468.
[4] Datta, Siddhant & Fard, Masoud & Chattopadhyay, Aditi. (2015). High-Speed Surfactant-Free Fabrication of Large Carbon Nanotube Membranes for Multifunctional Composites. Journal of Aerospace Engineering. 29. 04015060. 10.1061/(ASCE)AS.1943-5525.0000558.
[5] Li, Z., Liang, Z. Optimization of Buckypaper-enhanced Multifunctional Thermoplastic Composites. *Sci Rep* **7**, 42423 (2017).
[6] Jun Young Oh, Seung Jae Yang, Jun Young Park, Taehoon Kim, Kunsil Lee, Yern Seung Kim, Heung Nam Han, and Chong Rae Park "Easy Preparation of Self-Assembled High-Density Buckypaper with Enhanced Mechanical Properties" Nano Letters 2015 15 (1), 190-197
[7] Y.-C. Shin and S.-M. Kim, "Enhancement of the Interlaminar Fracture Toughness of a Carbon-Fiber-Reinforced Polymer Using Interleaved Carbon Nanotube Buckypaper" Applied Sciences, vol. 11, no. 15, p. 6821, Jul. 2021, doi: 10.3390/app11156821.
[8] M.R, Al-Hadrayi & Chwei, Zhou. (2016). Effect the stacking sequences of composite laminates under low velocity impact on failure modes by using carbon fiber reinforced polymer. 5. 2319-1813.
[9] MIRALON® :: Huntsman Corporation (HUN)
[10] Chandra, Y., Adhikari, S., Mukherjee, S. et al. "Unfolding the mechanical properties of buckypaper composites: nano- to macro-scale coupled atomistic-continuum simulations" Engineering with Computers 38, 5199–5229 (2022).
[11] A. Arteiro, G. Catalanotti, A.R. Melro, P. Linde, P.P. Camanho, "Micro-mechanical analysis of the effect of ply thickness on the transverse compressive strength of polymer composites" Composites Part A: Applied Science and Manufacturing, Volume 79, 2015
[12] Schaefer, Joseph D., Werner, Brian T., and Daniel, Isaac M. "Strain-rate-dependent Failure Criteria for Composite Laminates: Application of the Northwestern Failure Theory to Multiple Material Systems" United States: N. p., 2017. Web. doi:10.1007/978-3-319-63408-1_19
[13] https://energy.ornl.gov/CFCrush/materials/uou/8552_eu.pdf
[14] Yang, Y., Liu, X., Wang, Y. Q., Gao, H., Li, R., & Bao, Y. (2017). A progressive damage model for predicting damage evolution of laminated composites subjected to three-point bending. Composites Science and Technology, 151, 85-93.
[15] Microsoft Word - Hexcel_IM7_Material_Property_Data_Report_FINAL_4.22.11_Rev_A (wichita.edu)
[16] Shen, B., Liu, H., Lv, S., Li, Z., & Cheng, W. (2022). Progressive Failure Analysis of Laminated CFRP Composites under Three-Point Bending Load. Advances in Materials Science and Engineering, 2022.

[17] Venkatesan, K. R., Stoumbos, T., Inoyama, D., & Chattopadhyay, A. (2021). Computational analysis of failure mechanisms in composite sandwich space structures subject to cyclic thermal loading. Composite Structures, 256, 113086.
[18] Aboudi, J., Arnold, S. M., & Bednarcyk, B. A. (2013). Micromechanics of composite materials: a generalized multiscale analysis approach. Butterworth-Heinemann. https://doi.org/10.1016/B978-0-12-397035-0.00002-1
[19] ASTM D7264-16. (2016). Standard Test Method for Flexural Properties of Polymer Matrix Composite Materials. ASTM International, West Conshohocken, PA.
[20] Pang, Yutian, et al. "Data-driven trajectory prediction with weather uncertainties: A Bayesian deep learning approach." Transportation Research Part C: Emerging Technologies 130 (2021): 103326.
[21] Zhang, Xiaoge, and Sankaran Mahadevan. "Bayesian neural networks for flight trajectory prediction and safety assessment." Decision Support Systems 131 (2020): 113246.
[22] Hamza, Mohamed, et al. "Physics-based recurrent neural network model for flight path trajectory prediction under high-altitude stall." AIAA SCITECH 2023 Forum. 2023.
[23] Borkowski, L., C. Sorini, and A. Chattopadhyay. "Recurrent neural network-based multiaxial plasticity model with regularization for physics-informed constraints." Computers & Structures 258 (2022): 106678.
[24] Borkowski, L., T. Skinner, and A. Chattopadhyay. "Woven ceramic matrix composite surrogate model based on physics-informed recurrent neural network." Composite Structures 305 (2023): 116455.
[25] Ioffe, Sergey, and Christian Szegedy. "Batch normalization: Accelerating deep network training by reducing internal covariate shift." International conference on machine learning. pmlr, 2015.
[26] Kingma, Diederik P., and Jimmy Ba. "Adam: A method for stochastic optimization." arXiv preprint arXiv:1412.6980 (2014)